\documentclass{article}
\usepackage{spconf,amsmath,epsfig}
\usepackage[dvipsnames]{xcolor}
\usepackage{multirow}
\usepackage{graphicx}
\usepackage{multicol}
\usepackage{arydshln}
\usepackage{amsmath}
\usepackage{booktabs}
\usepackage{bbding}
\usepackage{textcomp}
\usepackage{tablefootnote}
\usepackage{threeparttable}
\usepackage{pifont}
\usepackage[accsupp]{axessibility}

\newcommand{\circled}[1]{\textcircled{\raisebox{-0.8pt}{#1}}}

\let\OLDthebibliography\thebibliography
\renewcommand\thebibliography[1]{
  \OLDthebibliography{#1}
  \setlength{\parskip}{0pt}
  \setlength{\itemsep}{0pt plus 0.3ex}
}

\begin{document}\sloppy

\def\x{{\mathbf x}}
\def\L{{\cal L}}
\def\F{\mathcal{F}}
\def\G{\mathcal{G}}
\definecolor{deepblue}{RGB}{32,56,100} 
\definecolor{deepred}{RGB}{132,60,12} 

\title{PAME: Self-Supervised Masked Autoencoder for No-Reference Point Cloud Quality Assessment}
\name{Ziyu Shan$^{\rm 1}$, Yujie Zhang$^{\rm 1}$, Qi Yang$^{\rm 2}$, Haichen Yang$^{\rm 1}$ Yiling Xu$^{\rm 1,\dagger}$ Shan Liu$^{\rm 2}$ \thanks{$^{\dagger}$Corresponding author: Yiling Xu.}\thanks{This paper is supported in part by National Natural Science Foundation of China (62371290, U20A20185) and 111 project (BP0719010). The corresponding author is Yiling Xu(e-mail: yl.xu@sjtu.edu.cn).}}
\address{$^{\rm 1}$Shanghai Jiao Tong University, $^{\rm 2}$Tencent\\ $^{\rm 1}$\{shanziyu, yujie19981026, yanghaichen, yl.xu\}@sjtu.edu.cn, $^{\rm 2}$\{chinoyang, shanl\}@tencent.com}

\maketitle

\begin{abstract}
No-reference point cloud quality assessment (NR-PCQA) aims to automatically predict the perceptual quality of point clouds without reference, which has achieved remarkable performance due to the utilization of deep learning-based models. However, these data-driven models suffer from the scarcity of labeled data and perform unsatisfactorily in cross-dataset evaluations. To address this problem, we propose a self-supervised pre-training framework using masked autoencoders (PAME) to help the model learn useful representations without labels. Specifically, after projecting point clouds into images, our PAME employs dual-branch autoencoders, reconstructing masked patches from distorted images into the original patches within reference and distorted images. In this manner, the two branches can separately learn content-aware features and distortion-aware features from the projected images. Furthermore, in the model fine-tuning stage, the learned content-aware features serve as a guide to fuse the point cloud quality features extracted from different perspectives. Extensive experiments show that our method outperforms the state-of-the-art NR-PCQA methods on popular benchmarks in terms of prediction accuracy and generalizability.
\end{abstract}
\section{Introduction}
\label{sec:intro}

With the improvement of 3D acquisition equipment, point clouds have become a prominent and popular media format in various scenarios, such as autonomous driving and virtual reality \cite{ wang2023improving,zhou2022blind}. A point cloud contains a large number of unordered points, each of which is described by its 3D coordinates and attributes (\textit{e.g.}, RGB color, normals). In practical applications, point clouds are subject to various distortions at each stage of their operational cycle (\textit{e.g.}, acquisition, compression, and transmission), thus affecting the perceptual quality of the human vision system (HVS). In order to improve the quality of experience, point cloud quality assessment (PCQA) has become an increasingly important research topic in both industry and academia. In this paper, we focus on NR-PCQA, which are designed for typical client applications where reference point clouds are not available. 

Reviewing the development of NR-PCQA, most methods are based on deep neural networks, which can be further classified into point-based, projection-based and multi-modal methods. The first category \cite{tliba2022representation, shan2023gpa, tliba2023pcqa, cheng2023no, xiong2023psi} directly infers point cloud quality based on the original 3D data and resorts to effective processing tools such as sparse convolution \cite{liu2023point} and graph convolution \cite{shan2023gpa}. In comparison, projection-based methods \cite{fan2022no, liu2021pqa, tao2021point, mu2023multi, yang2022no} project the point cloud into 2D images or videos and feed them into CNN-like backbones. Recently, some multi-modal methods \cite{zhang2022mm} have tried to extract independent features from 3D point cloud patches and 2D projected images and fuse the features from different modalities. Overall, although the above methods have presented remarkable performance when trained and tested on the same PCQA dataset, they still suffer from the scarcity of labeled data and perform unsatisfactorily in the cross-dataset evaluations. In fact, most PCQA datasets \cite{liu2023point,liu2022perceptual,yang2020predicting} provide only hundreds of samples with labels (\textit{i.e.,} mean opinion score, MOS) due to the high cost of annotation process, indicating that existing PCQA datasets are too small to train a learning-based model with good generalizability.

Due to the ability to utilize a large amount of unlabeled data, masked autoencoding is a potential solution to address the scarcity of labeled data for PCQA and boost the generalizability of NR-PCQA models. This technique has proven to be effective in various computer vision tasks \cite{he2022masked} as a popular self-supervised learning framework. In terms of classical masked autoencoding for images (\textit{i.e.,} masked image modeling, MIM), the autoencoder masks a portion of patches and encodes the visible patches, followed by reconstructing the masked patches to capture content-aware information. However, unlike the pre-trained models using classical MIM which mainly focus on high-level content-aware information, the NR-PCQA models should learn both high-level content-aware features representing point cloud content and low-level quality features capturing distortion information. 

\begin{figure*}[t]
    \centering
    \vspace{-0.6cm}
    \includegraphics[width=0.72\textwidth]{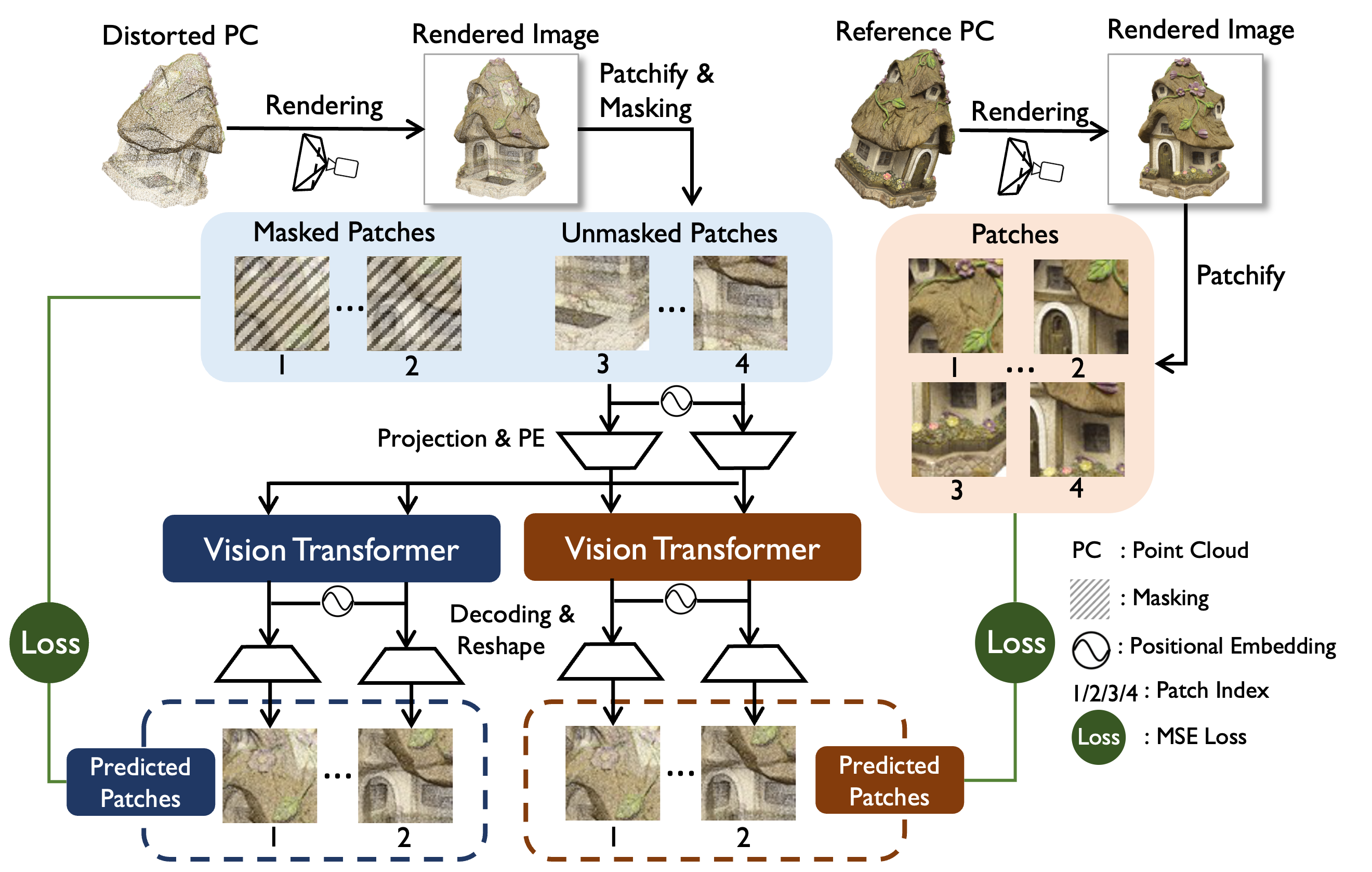}
    \caption{\textbf{Framework of the proposed pre-training method (PAME).} Unlabeled point cloud is first projected into images, and the images are then partitioned into patches and masked 50\%. Then the visible unmasked patches are embedded by convolutional layers along with positional embedding before being fed to two vision transformers. Subsequently, the encoded features are decoded to predict the masked patches and the corresponding patches of the projected images rendered from the reference point cloud. The content-aware and distortion-aware branch are color-marked in \bf\textcolor{deepblue}{BLUE} and \bf\textcolor{deepred}{RED}.}
    \label{fig:pipeline}
\end{figure*}

To achieve this goal, we propose a self-supervised pre-training framework for NR-\textbf{P}CQ\textbf{A} using \textbf{M}asking auto\textbf{E}ncoders (PAME). Our PAME employs a dual-branch structure, in which each branch is specialized in an autoencoder architecture, to separately learn high-level content-aware features (called content-aware branch) and low-level distortion-aware features (called distortion-aware branch) from projected images of distorted point clouds, as illustrated in Fig.\ref{fig:pipeline}. The distorted point cloud is first projected into images and partitioned into patches. Then about 50\% of these patches are masked and the unmasked patches are fed to the encoders (\textit{i.e.,} two vision transformers \cite{dosovitskiy2020image} sharing the same input) of both branches for encoding. Subsequently, for the distortion-aware branch, the encoded latent features are decoded to reconstruct the masked patches, in which manner the encoder can capture the pristine distortion distribution before masking. In comparison, the latent features of the content-aware branch are decoded to reconstruct the corresponding patches from the projected images of the reference point cloud, which focuses mainly on high-level content-aware information referring to  classical MIM. Following the self-supervised learning paradigm, we finally obtain two pre-trained encoders that are expected to extract content-aware features and distortion-aware features, which are then incorporated into our model and fine-tuned with labeled data.

In the fine-tuning stage, the labeled point cloud is rendered into multi-view images to mimic the subjective observation process; next the multi-view images are fed to the pre-trained encoders to generate content-aware and distortion-aware features. The content-aware features then serve as a guide to fuse distortion-aware features from different perspectives through a simple cross-attention mechanism, followed by a quality score regression module to predict objective scores. Extensive experiments show that our method exhibits superior prediction accuracy and generalization compared to the state-of-the-art NR-PCQA methods.

\section{Our Approach}
\label{sec:method}

\subsection{Overview}
\begin{figure}[t]
    \centering
    \vspace{-0.6cm}
    \includegraphics[width=0.25\textwidth]{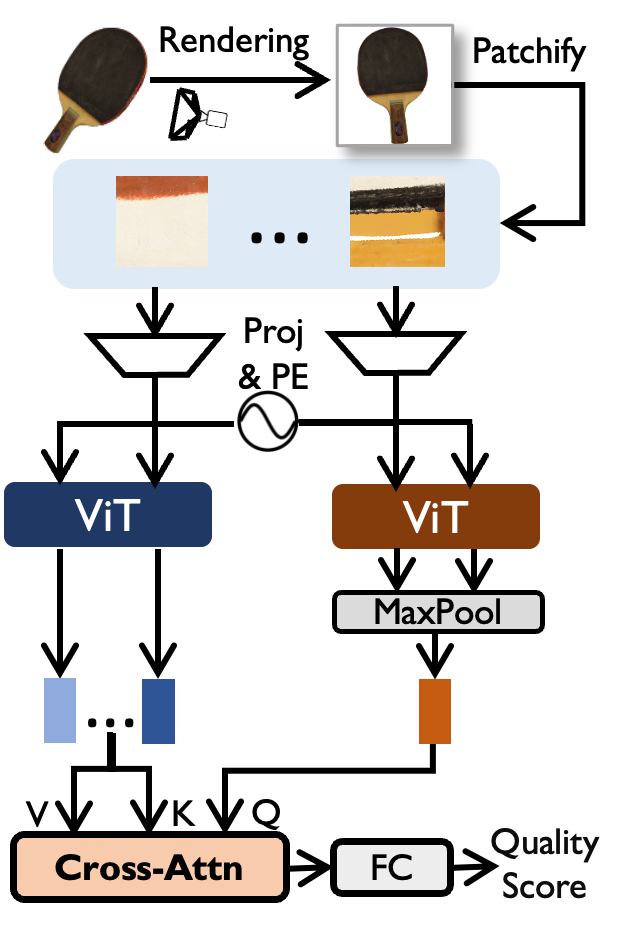}
    \caption{\textbf{Framework of the fine-tuning stage.} The labeled point cloud is rendered and patchfied, and the patches are then embedded. After encoded by $\mathcal{F}$ and $\mathcal{G}$, the features in content-aware branch are maxpooled to guide the fusion of distortion-aware features using the cross-attention mechanism.}
    \vspace{-0.3cm}
    \label{fig:finetune}
\end{figure}

Our method consists of a pre-training stage and a fine-tuning stage. 1) In the pre-training stage, our goal is to pre-train two encoders $\mathcal{F}$ and $\mathcal{G}$ (\textit{i.e.,} two ViT-B \cite{dosovitskiy2020image}) that can individually extract content-aware features and distortion-aware features. To achieve this, as illustrated in Fig.\ref{fig:pipeline}, the pre-training framework (PAME) is divided into a content-aware branch and a distortion-aware branch, and both branches are specialized in an autoencoder architecture. Concretely, both branches take the distorted point cloud as input and subsequently render it into images, followed by partitioning the images into patches. Then about 50\% of the patches are masked, while the unmasked patches are embedded into features by a linear projection together with positional embedding following \cite{he2022masked}. Subsequently, the embedded features are fed to the vision transformers and decoded to predict the masked patches and the corresponding patches of the projected image rendered from the reference point cloud, respectively. 2) In the fine-tuning stage, as illustrated in Fig.\ref{fig:finetune}, the labeled point cloud is rendered into multi-view images and fed to the pre-trained $\mathcal{F}$ and $\mathcal{G}$. Then the extracted content-aware features serve as a guide to integrate the distortion-aware features from different perspectives through a simple cross-attention mechanism, followed by fully-connected layers to regress the quality score.

\subsection{Pre-training via Masked Autoencoding}
To begin with, we define a distorted point cloud $X = \{ {x}_1 , {x}_2 \cdots ,{x}_{N}\} \in \mathbb R ^ {N \times 6}$, where each point ${x}_i$ consists of its 3D coordinates and RGB attributes. Then, we define the corresponding reference point cloud $Y \in \mathbb R ^ {M \times 6}$. Now our goal is to render $X$ and $Y$ into images and pre-train encoders $\mathcal{F}$ and $\mathcal{G}$ via masked autoencoding.

\noindent\textbf{Multi-view Rendering.} Instead of directly performing masked autoencoding in 3D space, we render point clouds into 2D images to achieve pixel-to-pixel correspondence between the rendered images of $X$ and $Y$,  which facilitates the computation of pixel-wise reconstruction loss between the predicted patches and the ground truth patches in the content-aware branch.

To perform the rendering, we translate $X$ (or $Y$) to the origin and geometrically normalize it to the unit sphere to achieve a consistent spatial scale. Then, to fully capture the quality information of 3D point clouds, we render $X$ into images from 12 viewpoints that are evenly distributed in 3D space at a fixed viewing distance. For the sake simplicity,  we denote a randomly selected rendered image of $X$ as $I_X \in \mathbb R^{H\times W \times 3}$ and $I_Y$ has a similar meaning.

\noindent\textbf{Patchify and Masking.} After obtaining the rendered image $I_X$, we partition it into non-overlapping $16 \times 16$ patches following \cite{dosovitskiy2020image}. Then we randomly sample a subset of patches and mask the remaining ones, where the masking ratio is relaxed to 50\% instead of the high ratio in \cite{he2022masked} (\textit{e.g.,} 75\% and even higher) because some point cloud samples in PCQA datasets exhibit severe distortions, necessitating more patches to extract effective quality and content-aware information. In addition, the relatively low masking ratio can mitigate the effect of the background of $I_X$.

\noindent\textbf{Patch Embedding.} To embed the unmasked patches in the feature space, we flatten the patch to a 1D vector for linear projection with added positional embedding, following \cite{dosovitskiy2020image}.  Given an unmasked patch ${p_X}$, the embedding process can be formulated as:
\begin{equation}
    f_X = \mathrm{MLP}(\mathrm{Flatten}(p_X))+\mathrm{PE},
\end{equation}
where $f_X$ is the embedded feature, $\mathrm{MLP}$ and $\mathrm{PE}$ are linear layers and positional embedding, respectively.

\noindent\textbf{Encoding by Vision Transformers.} The embedded feature is fed to $\mathcal{F}$ and $\mathcal{G}$ for encoding in both content-aware and distortion-aware branches, which both consist of $L$ consecutive identical transformer blocks. Taking the $l$-th block {in $\mathcal{F}$} as an example, it takes the output $f_{X,\F}^{l-1}$ of the previous $(l-1)$-th block as input to compute its output $f_{X,\F}^l$:
\begin{equation}
\begin{split}
   & \tilde{f}_{X,\F}^l = \mathrm{MSA}(\mathrm{LN}(f_{X,\F}^{l-1})) + f_{X,\F}^{l-1},\\
   & f_{X,\F}^l = \mathrm{MLP}(\mathrm{LN}(\tilde{f}_{X,\F}^l))+\tilde{f}_{X,\F}^l,
\end{split}
\label{eq:encode}
\end{equation}
where $\mathrm{MSA}$ and $\mathrm{LN}$ denotes multi-headed self-attention operation and layer normalization. {$f_{X,\G}^l$ in $\G$ can be obtained using the same paradigm. }

Considering the non-trivial characteristic of predicting the pristine patches from neighboring distorted patches in the content-aware branch, the vision transformer in this branch is initialized with the weight  optimized by performing MIM on ImageNet-1K \cite{he2022masked} to extract content-aware information more effectively.

\noindent\textbf{Decoding and Reconstruction Loss.} After obtaining the encoded features $f_{X,\F}^L$ and $f_{X,\G}^L$, we combine them with the positional embedding of all the patches of $I_X$, which makes the decoder aware of the locations of the masked patches. This combination is then separately fed to two shallow decoders consisting of $L'$ transformer blocks to generate $f^{L+L'}_{X,\F}$ and $f^{L+L'}_{X,\G}$. Finally, we use one-layer MLP for projection and reshape the projected feature to $16 \times 16$ to form the predicted patches $\hat{p}_X$ and $\hat p_Y$, which can be formulated as:
\begin{equation}
\begin{aligned}
    \hat{p}_X &= \mathrm{Reshape}(\mathrm{MLP}(f^{L+L'}_{X,\F})),\\
    \hat{p}_Y &= \mathrm{Reshape}(\mathrm{MLP}(f^{L+L'}_{X,\G})).
\end{aligned}
\end{equation}

We denote the predicted patches combined with the unmasked patches as $\hat I_X$ and $\hat I_Y$, and the reconstruction loss for the distortion-aware branch and content-aware branch can be formulated as:
\begin{equation}
    \mathcal{L}_{rec} = \alpha \big\Vert \hat I_X - I_X \big\Vert_2^2 + (1-\alpha) \big\Vert \hat I_Y - I_Y \big\Vert_2^2,
\end{equation}
where $\alpha$ is a weighting coefficient. Note that the decoder and reconstruction loss are only used during pre-training.

\subsection{Fine-tuning with Labeled Data}
After obtaining the pre-trained encoders $\mathcal{F}$ and $\mathcal{G}$, we next fine-tune it with labeled point clouds. For a labeled point cloud $Z$, we render it into images from six perpendicular viewpoints (\textit{i.e.,} along the positive and negative directions of $x,y,z$ axes). The rendered images are then patchfied, embedded and encoded identically to pre-training to generate the multi-view features.

To fuse the multi-view features, we first maxpool the extracted features of the content-aware branch to form the most prominent content-aware feature.
Then we fuse the multi-view distortion-aware features through a simple multi-headed cross-attention mechanism $\mathrm{MCA}$:
\begin{equation}
     F = \mathrm{MCA} (\mathrm{MaxPool}(\{f^{L,i}_{Z,\G}\}_{i=1}^6), \{f^{L,i}_{Z,\F}\}_{i=1}^6, \{f^{L,i}_{Z,\F}\}_{i=1}^6)
\label{eq:fusion}
\end{equation}
where $ F$ is the fused feature, $\{f^{L,i}_{Z,\F}\}_{i=1}^6$ (or $\{f^{L,i}_{Z,\G}\}_{i=1}^6$) is the multi-view encoded features from six perspectives. In Eq.\eqref{eq:fusion}, intuitively, the content-aware feature acts as a query and computes similarity with each $f^{L,i}_{Z,\F}$ to find the content-awareally active viewpoints and fuse their distortion-aware features accordingly.

It is worth noting that only distorted point clouds are used during the fine-tune stage and the final inference, and the MOSs of reference point clouds are not available during pre-training, thus our method still belongs to NR-PCQA, as similarly did in \cite{lin2018hallucinated}.

\vspace{-0.3cm}
\subsection{Quality Regression and Loss Function}
After multi-view fusion, the final feature $ F$ is fed to two-fold fully-connected layers to regress the predicted quality score $\hat q$. Inspired by \cite{zhang2022mm}, our loss function includes two parts: mean square error (MSE) and rank error. The MSE optimizes the model to improve the prediction accuracy, which is formulated as:
\begin{equation}
    \mathcal{L}_{mse} = \frac{1}{B} \sum_{i=1}^B (\hat q_i - q_i)^2
\end{equation}
where $\hat q_i$ is the predicted quality score of $i$-th sample in a mini-batch with the size of $B$, and $q_i$ is the corresponding ground truth MOS.

To better recognize quality differences for the point clouds with close MOSs, we use a differential ranking loss \cite{sun2022deep} to model the ranking relationship between $\hat q$ and $q$:
\vspace{-0.3cm}
\begin{equation}
\begin{aligned}
    \mathcal{L}_{rank} = \frac{1}{B^2} \sum_{i=1}^B & \sum_{j=1}^B \!\max\! \left(0,\left|q_{i}-q_{j}\right|\!-\!e\left(q_{i}, q_{j}\right)\! \cdot\! \left(\hat{q}_{i}-\hat{q}_{j}\right)\right), \\
    &e\left(q_{i}, q_{j}\right)=\left\{\begin{array}{r}
1, q_{i} \geq q_{j} \\
-1 , q_{i} < q_{j}
\end{array}\right.
\end{aligned}
\end{equation}

\begin{equation}
    \mathcal{L}_{fine} = \beta \mathcal{L}_{mse} + (1-\beta) \mathcal{L}_{rank}
\end{equation}
where the hyper-parameter $\beta$ is to balance the two losses.

\begin{table*}[t]\small
\centering 
\vspace{-0.6cm}
\renewcommand\arraystretch{0.9}
\caption{Performance results on the LS-PCQA \cite{liu2023point}, SJTU-PCQA \cite{yang2020predicting} and WPC \cite{liu2022perceptual} databases. ``P" and ``I" stand for the method is based on the point cloud and image modality, respectively. ``$\uparrow$"/``$\downarrow$" indicates that larger/smaller is better. Best and second performance results are marked in {\bf\textcolor{red}{RED}} and {\bf\bf\textcolor{blue}{BLUE}} for FR-PCQA and NR-PCQA methods. ``FT'' indicates fine-tuning.}
\begin{tabular}{c:c:c|ccc|ccc|ccc}
\toprule
    \multirow{2}{*}{Ref}&\multirow{2}{*}{Modal}&\multirow{2}{*}{Methods} & \multicolumn{3}{c|}{LS-PCQA \cite{liu2023point}} & \multicolumn{3}{c|}{SJTU-PCQA \cite{yang2020predicting}} & \multicolumn{3}{c}{WPC \cite{liu2022perceptual}} \\ \cline{4-12}
        & & & SROCC$\uparrow$      & PLCC$\uparrow$        & RMSE $\downarrow$  & SROCC$\uparrow$      & PLCC$\uparrow$        & RMSE $\downarrow$    & SROCC$\uparrow$      & PLCC$\uparrow$          & RMSE $\downarrow$ \\ \hline
\multirow{6}{*}{FR} 
 &P&PSNR-yuv  & \bf\textcolor{blue}{0.548} & \bf\textcolor{blue}{0.547} & 0.155 & 0.704 & 0.715 & 0.165 & 0.563 & 0.579 & 0.186 \\
  &P&PointSSIM & 0.180 & 0.178 & 0.183 & 0.735 & 0.747 & 0.157 & 0.453 & 0.481 & 0.200 \\
 &P&PCQM      & 0.439 & 0.510 & \bf\textcolor{blue}{0.152} & 0.864 & 0.883 & \bf\textcolor{blue}{0.112} & \bf\textcolor{red}{0.750} & \bf\textcolor{red}{0.754} & \bf\textcolor{red}{0.150} \\
 &P&GraphSIM  & 0.320 & 0.281 & 0.178 & 0.856 & 0.874 & 0.114 & 0.679 & 0.693 & 0.165 \\
&P&MS-GraphSIM & 0.389 & 0.348 & 0.174 & \bf\textcolor{blue}{0.888} & \bf\textcolor{blue}{0.914} & \bf\textcolor{red}{0.096} & \bf\textcolor{blue}{0.704} & \bf\textcolor{blue}{0.718} & \bf\textcolor{blue}{0.159} \\

 &P&MPED  & \bf\textcolor{red}{0.659} & \bf\textcolor{red}{0.671} & \bf\textcolor{red}{0.138} & \bf\textcolor{red}{0.898} & \bf\textcolor{red}{0.915} & \bf\textcolor{red}{0.096} & 0.656 & 0.670  & 0.169
 \\\hdashline
\multirow{6}{*}{NR} 
 &I&PQA-Net   & 0.588 & 0.592 & 0.202 & 0.659 & 0.687 & 0.172 & 0.547 & 0.579 & 0.189\\
 &I&IT-PCQA   & 0.326 & 0.347 & 0.224 & 0.539 & 0.629 & 0.218 & 0.422 & 0.468 & 0.221 \\
 &P&GPA-Net   & 0.592 & 0.619 & 0.186 & \bf\textcolor{blue}{0.878} & 0.886 & 0.122 & 0.758 & 0.769 & 0.162 \\
 &P&ResSCNN   & \bf\textcolor{blue}{0.594} & \bf\textcolor{blue}{0.624} & \bf\textcolor{blue}{0.172} & 0.834 & 0.863 & 0.153 & 0.735 & 0.752 & 0.177 \\
 &P+I&MM-PCQA   & 0.581 & 0.597 & 0.189 & 0.876 & \bf\textcolor{blue}{0.898} & \bf\textcolor{blue}{0.109} & \bf\textcolor{blue}{0.761} & \bf\textcolor{blue}{0.774} & \bf\textcolor{blue}{0.149} \\
 &P&\textbf{PAME+FT} & \bf\textcolor{red}{0.608}& \bf\textcolor{red}{0.630} & \bf\textcolor{red}{0.167}  & \bf\textcolor{red}{0.889} & \bf\textcolor{red}{0.905} &
 \bf\textcolor{red}{0.101} & \bf\textcolor{red}{0.763} & \bf\textcolor{red}{0.781} &  \bf\textcolor{red}{0.145} \\

\bottomrule
\end{tabular}
\label{tab:1}
\end{table*}

\section{Experiments}
\label{sec:experiment}
\subsection{Datasets and Evaluation Criteria}
\noindent\textbf{Datasets.} Our experiments are based on three PCQA datasets, including LS-PCQA \cite{liu2023point}, SJTU-PCQA \cite{yang2020predicting}, and WPC \cite{liu2022perceptual}. The pre-training is conducted on the complete LS-PCQA, which is a large-scale PCQA dataset and contains 104 reference point clouds and 24,024 distorted point clouds, and each reference point cloud is impaired with 33 types of distortion under 7 levels. In the fine-tuning stage, the model is trained on all three datasets separately using labeled data, where SJTU-PCQA includes 9 reference point clouds and 378 distorted samples impaired with 7 types of distortion under 6 levels, while WPC contains 20 reference point clouds and 740 distorted sampled degraded by 5 types of distortion.

\noindent\textbf{Evaluation Metrics.} Three widely adopted evaluation metrics are employed to quantify the level of agreement between predicted quality scores and MOSs: Spearman rank order correlation coefficient (SROCC), Pearson linear correlation coefficient (PLCC), and root mean square error (RMSE). To ensure consistency between the value ranges of the predicted scores and subjective values, nonlinear Logistic-4 regression is used to align their ranges following \cite{shan2023gpa}.

\subsection{Implementation Details}
Our experiments are performed using PyTorch on NVIDIA 3090 GPUs. All point clouds are rendered into images with a resolution of $512\times512$ and cropped to $224\times 224$ following \cite{zhang2022mm}. The encoders $\mathcal{F}$ and $\mathcal{G}$ are ViT-B \cite{dosovitskiy2020image}.

\noindent\textbf{Pre-training.} The pre-training is performed for 200 epochs, the initial learning rate is 3e-4, and the batch size is 16 by default. 
Adam optimizer is employed with weight decay of 0.0001. Besides, $\alpha$ is set to 0.7.

\noindent\textbf{Fine-tuning.} The fine-tuning is performed for 20 epochs for LS-PCQA, while 150 epochs for SJTU-PCQA and WPC. The initial learning rate is 0.003, and the batch size is 16 by default. The multi-headed cross-attention employs 8 heads and $d_f$ is empirically set to 1024. Besides, $\beta$ is set to 0.5.

\noindent\textbf{Dataset Split.} Considering the limited dataset scale, in fine-tuning stage, 5-fold cross validation is used for SJTU-PCQA and WPC to reduce content bias. Take SJTU-PCQA for example, in each fold, the dataset is split into train-test with ratio 7:2 according to the reference point clouds, where the performance on testing set with minimal training loss is recorded and then averaged across five folds to get the final result. A similar procedure is repeated for WPC where the train-test ratio is 4:1. As for the large-scale LS-PCQA, it is split into train-val-test with a ratio around 8:1:1 (no content overlap exists). The result on the testing set with the best validation performance is recorded. Note that the pre-training is only conducted on the training set of LS-PCQA.

\subsection{Comparison with State-of-the-art Methods}
11 state-of-the-art PCQA methods are selected for comparison, including 6 FR-PCQA and 5 NR-PCQA methods. The FR-PCQA methods include PSNR-yuv \cite{torlig2018novel}, PointSSIM \cite{alexiou2020pointssim}, PCQM \cite{meynet2020pcqm}, GraphSIM \cite{yang2020inferring}, MS-GraphSIM \cite{zhang2021ms}, and MPED \cite{yang2022mped}. The NR-PCQA methods include PQA-Net \cite{liu2021pqa}, IT-PCQA \cite{yang2022no}, GPA-Net \cite{shan2023gpa}, ResSCNN \cite{liu2023point}, and MM-PCQA \cite{zhang2022mm}. We conduct experiments to test the prediction accuracy following the cross-validation configuration. 

\begin{table}[t]
\renewcommand\tabcolsep{3.3pt}
\centering
\vspace{-0.6cm}
\caption{Cross-dataset evaluation for NR-PCQA methods. Training and testing are both conducted on complete datasets. Results of PLCC are reported.}
\scriptsize
\begin{tabular}{cc|ccccc}
    \toprule  
    Train & Test & PQA-Net & GPA-Net & ResSCNN & MM-PCQA & PAME+FT \\ 
    \midrule  
    LS & SJTU & 0.342 & 0.556 & 0.546 & \bf\textcolor{blue}{0.581} & \bf\textcolor{red}{0.632}\\
    LS & WPC & 0.266 & 0.433 & \bf\textcolor{blue}{0.466} & 0.454 &\bf\textcolor{red}{0.499} \\
    WPC & SJTU & 0.235 & 0.553 & 0.572 & \bf\textcolor{blue}{0.612} & \bf\textcolor{red}{0.628} \\
    SJTU & WPC & 0.220 & \bf\textcolor{blue}{0.418} & 0.269 & 0.409 & \bf\textcolor{red}{0.477} \\
		\bottomrule  
    \end{tabular}
    \vspace{-0.3cm}
    \label{tab:cross}
\end{table}

\begin{figure}[t]
    \centering
    
  \includegraphics[width=0.33\textwidth]{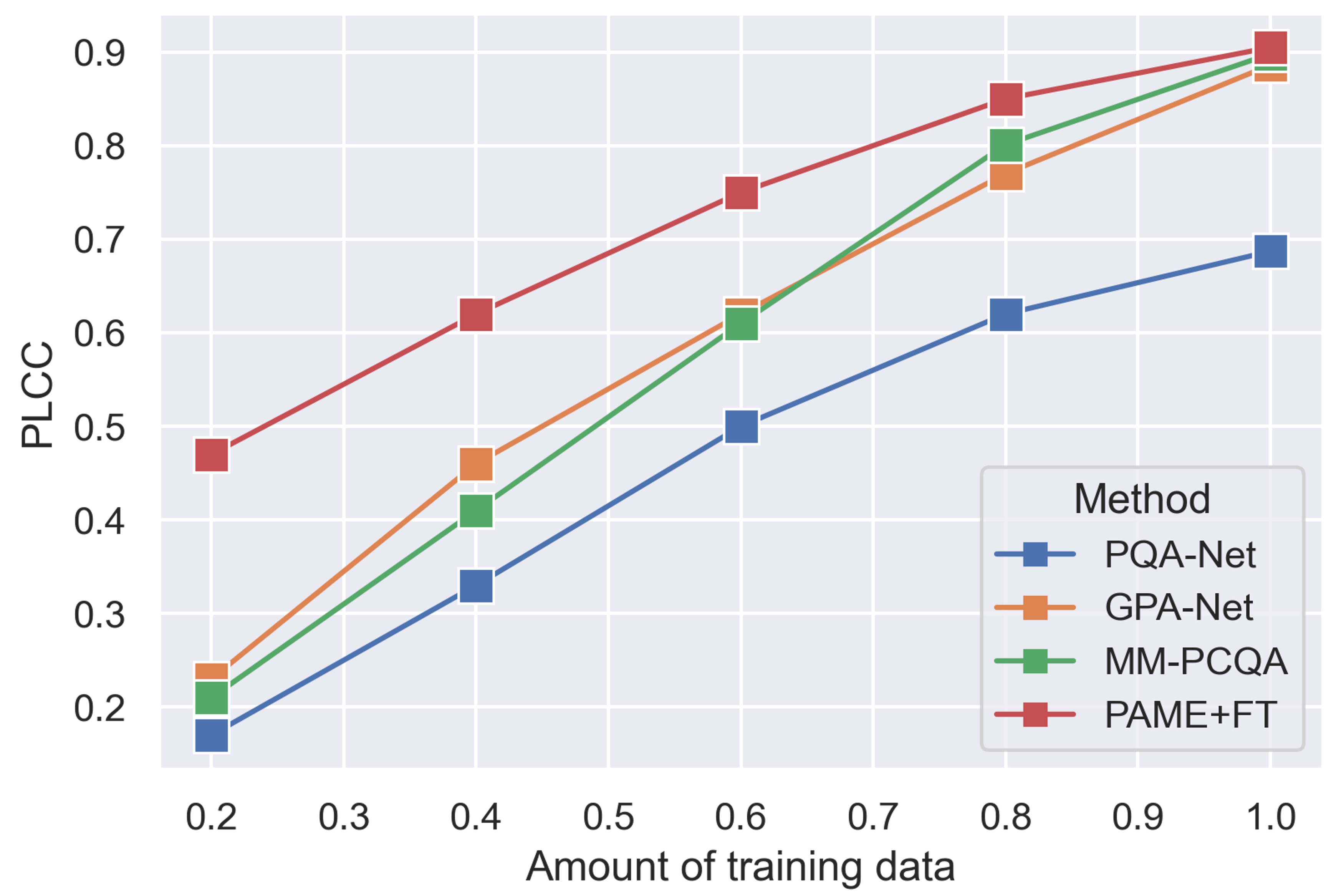}
  \vspace{-0.3cm}
    \caption{PLCCs of the NR-PCQA methods with less labeled data on SJTU-PCQA.}
    \vspace{-0.6cm}
    \label{fig:few_shot}
\end{figure}

\begin{table}[t]
\vspace{-0.3cm}
\renewcommand\arraystretch{1.0}
    \centering
    \caption{Ablation on the components of our model on SJTU-PCQA. `\ding{51}' or `\ding{55}' means the setting is preserved or discarded. `Con.' and `Dis.' indicates content-aware and distortion-aware branch.}
    \label{tab:components}
    \setlength{\tabcolsep}{2.5pt}
    \scriptsize
    \begin{tabular}{c|ccc|cc}
    \toprule
    Index & Pre-training & Multi-view fusion \tnote{1} & Fine-tuning loss \tnote{1} & SROCC & PLCC\\ \midrule
    \circled{1}&\ding{51} & \ding{51} & \ding{51} & \bf\textcolor{red}{0.889} & \bf\textcolor{red}{0.905} \\ \hdashline
    \circled{2}&\ding{55} & \ding{51} & \ding{51} & 0.724 & 0.772 \\
    \circled{3}& Con. only & \ding{51} & \ding{51} & 0.854 & 0.867 \\
    \circled{4}& Dis. only & \ding{51} & \ding{51}  & 0.852  & 0.901 \\ \hdashline
    \circled{5} & \ding{51} & MaxPool+Concat  & \ding{51} & 0.835 & 0.854 \\
    \circled{6} & \ding{51}  & \ding{51} & MSE only & \bf\textcolor{blue}{0.885} & \bf\textcolor{blue}{0.899} \\ 
    \bottomrule
    \end{tabular}
    \vspace{-0.3cm}
\end{table}
From Tab.\ref{tab:1} we can see that our method outperforms the NR-PCQA methods across three datasets. Furthermore, we perform cross-dataset evaluation and compare the performances with less labeled data to test the generalizability. The results are in Tab.\ref{tab:cross} and Fig.\ref{fig:few_shot}, from which we can conclude that our method yields robust performance across different datasets and even with less labeled data.

\subsection{Ablation Studies}
In Tab. \ref{tab:components}, we report the results on SJTU-PCQA with the conditions of removing some components. From Tab.\ref{tab:components}, we have the following observations: 1) Seeing \circled{1} and \circled{2} - \circled{4}, the pre-training framework brings the most significant improvements when preserving both content-aware and distortion-aware branches. 2) Seeing \circled{1} and \circled{5} - \circled{6}, the multi-view fusion method performs better than simply maxpooling and concatenation. 3) Seeing \circled{1} and \circled{7}, the performance is close, which demonstrates the robustness of our model using different fine-tuning loss functions.

\section{Conclusion}
In this paper, we propose a novel no-reference point cloud quality method based on masked autoencoders. Specializing in a dual-branch structure, the proposed PAME learns content-aware and distortion-aware features through individually masked autoencoding. Furthermore, in the fine-tuning stage, we propose the multi-view fusion to attentively integrate the distortion-aware features from different perspectives. Experiments show that our method presents competitive and generalizable performance.

\bibliographystyle{IEEEbib}
\bibliography{icme2023template}

\end{document}